\documentclass[lettersize,journal]{IEEEtran}
\usepackage{amsmath,amsfonts}
\usepackage{algorithmic}
\usepackage{array}
\usepackage[caption=false,font=normalsize,labelfont=sf,textfont=sf]{subfig}
\usepackage{textcomp}
\usepackage{stfloats}
\usepackage{url}
\usepackage{verbatim}
\usepackage{graphicx}
\usepackage[numbers]{natbib}
\usepackage{multicol}
\usepackage[bookmarks=true]{hyperref}
\usepackage{mathtools}
\usepackage{graphicx}
\usepackage{multirow}
\usepackage{caption}
\newtheorem{remark}{Remark}
\usepackage{balance}
\DeclareRobustCommand*{\IEEEauthorrefmark}[1]{\raisebox{0pt}[0pt][0pt]{\textsuperscript{\footnotesize #1}}}

\begin{document}
\title{Contact-Aware Motion Planning \\Among Movable Objects}
\author{
	\IEEEauthorblockN{Haokun Wang\IEEEauthorrefmark{1}}, 
	\IEEEauthorblockN{Qianhao Wang\IEEEauthorrefmark{2}}, 
	\IEEEauthorblockN{Fei Gao\IEEEauthorrefmark{2*}},
	\IEEEauthorblockN{Shaojie Shen\IEEEauthorrefmark{1*}} 
	\thanks{Corresponding authors: Fei Gao and Shaojie Shen.}
	\thanks{\IEEEauthorrefmark{1}H. Wang and S. Shen are with Cheng Kar-Shun Robotics Institute, The Hong Kong University of Science and Technology, Hong Kong SAR, China (e-mail: hwangeh@connect.ust.hk; eeshaojie@ust.hk).}
	\thanks{\IEEEauthorrefmark{2}Q. Wang and F. Gao are with the Institute of Cyber-Systems and Control, College of Control Science and Engineering, Zhejiang University, Hangzhou 310027, China, and with the Huzhou Institute, Zhejiang University, Huzhou 313000, China (e-mail:qhwangaa@zju.edu.cn; fgaoaa@zju.edu.cn).}
}


\maketitle

\begin{abstract}
Most existing methods for motion planning of mobile robots involve generating collision-free trajectories. However, these methods focusing solely on contact avoidance may limit the robots' locomotion and can not be applied to tasks where contact is inevitable or intentional.
To address these issues, we propose a novel contact-aware motion planning (CAMP) paradigm for robotic systems. 
Our approach incorporates contact between robots and movable objects as complementarity constraints in optimization-based trajectory planning. 
By leveraging augmented Lagrangian methods (ALMs), we efficiently solve the optimization problem with complementarity constraints, producing spatial-temporal optimal trajectories of the robots.
Simulations demonstrate that, compared to the state-of-the-art method, our proposed CAMP method expands the reachable space of mobile robots, resulting in a significant improvement in the success rate of two types of fundamental tasks: navigation among movable objects (NAMO) and rearrangement of movable objects (RAMO). Real-world experiments show that the trajectories generated by our proposed method are feasible and quickly deployed in different tasks.
\end{abstract}


\section{Introduction}
\IEEEPARstart{I}{nteraction} with movable objects in an environment is a natural and essential human behavior. For instance, when navigating through a narrow space to reach the other side of a room, humans do not go around all objects in their path—instead, they selectively push chairs or tables that obstruct their way. Additionally, in exploring unknown environments, humans can actively open a movable window or door, thus exploring previously inaccessible areas. 

\begin{figure}[t]
	\centering
	\includegraphics[width=0.5\textwidth]{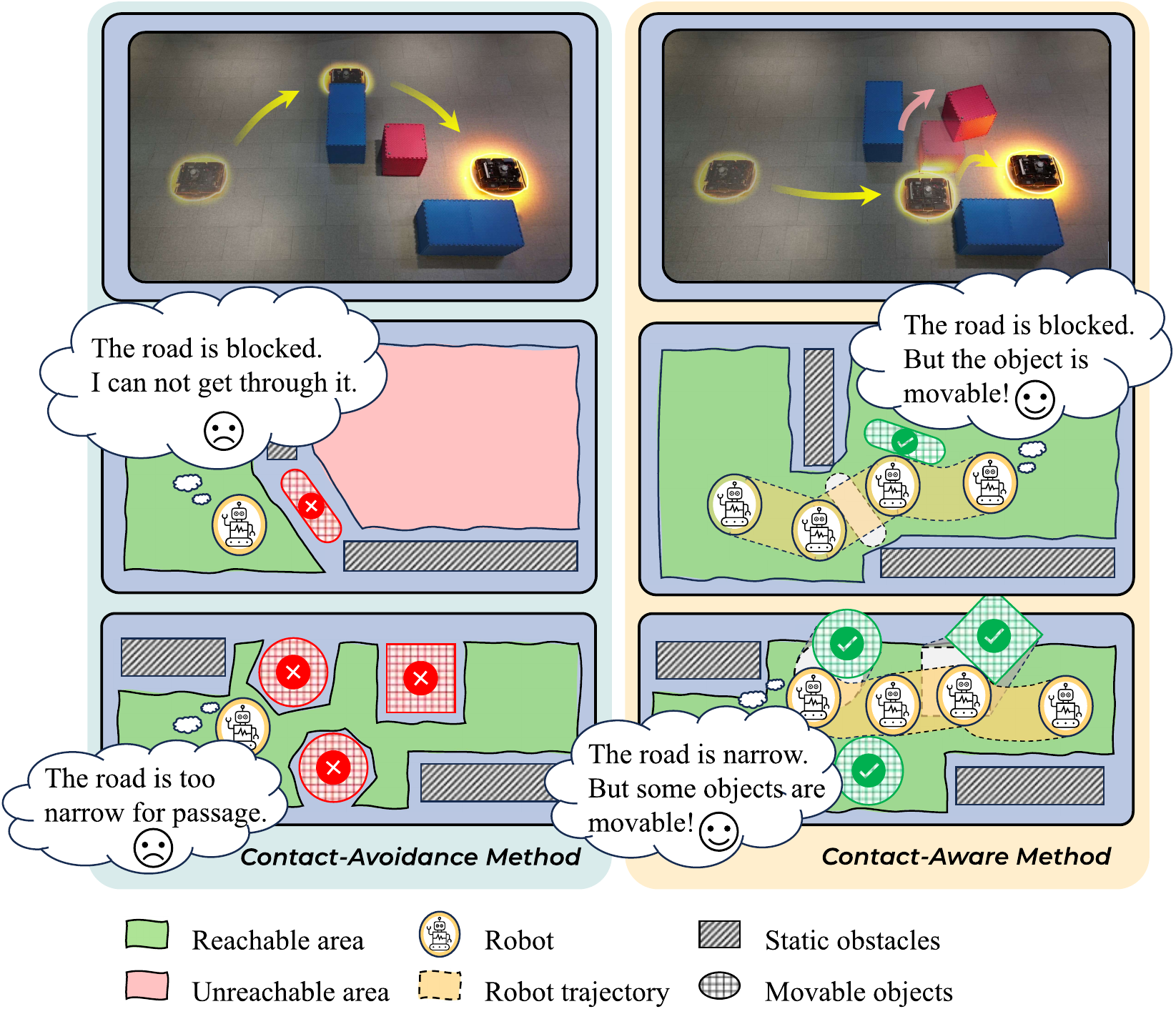}
	\caption{Compare to contact-avoidance method, contact-aware motion planning allows a mobile robot to push movable objects actively in crowded environments.}
	\label{fig-introduction}
\end{figure}

For physical robotic systems, most existing planning methods are insufficient to support the ability to interact with objects like humans in a complex environment.
These contact-avoidance methods treat all objects in the scenario as static obstacles and always generate collision-free trajectories for the robot \cite{wang2022gcopter, tobia2023gcs, jesus2022faster, zhou2019fastplaner}. While this strategy maximizes the safety of robot movement, it reduces the reachable space for the robot, limiting its mobility (as shown in Fig.\ref{fig-introduction}). In many tasks, such as navigating in a crowded room and pushing doors, it is inevitable or even necessary for the robot to make contact with movable objects in the environment. To efficiently accomplish these tasks and acquire locomotion capabilities like humans, the robot must possess contact-aware motion planning ability while preserving rational obstacle avoidance behaviors.

However, the existing research on contact-aware motion planning algorithms for robots is inadequate, especially in exploring methods to obtain feasible trajectories for continuous, smooth, and long-distance contact. 
Pioneering work by Stilman et al. \cite{stilman2005navigation} employed searching methods in a discrete action space to generate a sequence of primitive actions. While this high-level action planning effectively guides the robot to manipulate objects, ensuring the feasibility of trajectories between adjacent discrete states remains challenging, particularly during make and break contact. 
Additionally, the feasibility of contacts in the trajectories is crucial for the precise execution of practical tasks by the robots, such as wheeled balancing robots \cite{vicotr2023nonsmoothcontact} and bipedal robots \cite{posa2014direct}.

In this work, we propose a novel optimization-based contact-aware motion planning (CAMP) paradigm. CAMP leverages the information of movable objects in the environment to produce feasible trajectories for a mobile robot that involve predictable contact. The contacts between the robot and movable objects are embedded into the trajectory optimization problem as complementarity constraints. We efficiently solve this complex problem using the augmented Lagrangian method (ALM) \cite{hestenes1969multiplier,powell1969method,rockafellar1974augmented,birgin2014practical}.

By comparing our proposed method with classic motion planning methods in two fundamental planning tasks (definitions are detailed in Section \ref{sec-related_contact_traj}), the navigation among movable objects (NAMO) and the rearrangement of movable objects (RAMO), we demonstrate the efficiency of CAMP in solving the optimization problem and success rate of tasks accomplishment. Simulation and experiments fully validate the feasibility of generated trajectories. Furthermore, we open-source the code of CAMP, aiming to foster more attention from the community towards this approach. 

In summary, our main contributions are as follows.
\begin{itemize}
	\item We propose a novel contact-aware motion planning framework for mobile robots, allowing the robots to actively and predictably contact movable objects.
	\item Various simulations in two types of fundamental tasks, NAMO and RAMO, demonstrate the great potential of our method to extend the locomotion of mobile robots.
	\item Sufficient real-world experiments validate that our method can produce viable long-duration, long-distance contact trajectories and can be quickly deployed in different tasks.
	\item We will open-source our code to facilitate the community's expansion of related work.
\end{itemize}

\section{Related Work}
In this section, we review the relevant research from two perspectives: the contact-aware motion planning approaches for robots and the essential feasible trajectory optimization methods. In Section
\ref{sec-related_contact_traj}, we primarily discuss the significance of contact in various applications and the challenges that arise in the planning field due to the lack of contact consideration. Regarding feasible trajectory optimization (see Section \ref{sec-related_feasible_traj}), we highlight how overly conservative feasibility criteria can limit the motion capabilities of mobile robots.
\subsection{Contact-Aware Trajectory Planning for Robots}
\label{sec-related_contact_traj}
As figured in Fig.\ref{fig-introduction}, contact-aware motion planning methods can develop the locomotion of mobile robots and accomplish tasks that conservative planning methods cannot. Two categories of tasks are particularly essential: navigation among movable objects (NAMO) and rearrangement of movable objects (RAMO). 
On the one hand, compared to basic navigation tasks, NAMO tasks require mobile robots to fully utilize the interactable information of objects in the scene to complete the navigation efficiently. 
On the other hand, the RAMO skill specifies a desired target state for the movable objects. The main difference between RAMO and NAMO is that the target state of the movable objects can be arbitrary for the latter but not for the former. NAMO and RAMO are essential because they are fundamental sub-tasks for constructing more complex contact-rich tasks, such as exploring an unknown environment, 
completing daily chores in an open space, 
and so on. Contact-aware motion planning approaches must balance the cost of contact with movable objects and the cost of avoiding them to generate reasonable trajectories.

For these two types of tasks, much of the relevant research has focused on allocating reasonable discrete action sequences and determining appropriate target states \cite{saxena2023planning,saxena2023mamo,levihn2013hierarchical}. Planning strategies based on searching or sampling methods often face the challenge called the curse of dimensionality, as each state node generates two different motion modes due to whether contact occurrence \cite{russ2023underactuated,posa2014direct}. Additionally, sparse action sequences can amplify the potential collision risks between the robot and objects. Feasible continuous reference trajectories can improve the success rate of contact-rich tasks. 

In the research on motion planning of robotic systems with multiple motion modes and control strategies for bipedal robots, the employment of contact-aware methods is no longer a secret. Posa et al. \cite{posa2014direct} proposed a direct method that can produce continuous trajectories with contact properties for robots without predefined motion modes. In a recent study focused on wheeled balancing robots \cite{vicotr2023nonsmoothcontact}, contact switches and impacts are incorporated into the trajectory optimization to adapt to discontinuous terrains. 

Furthermore, there are data-driven approaches \cite{bianchini2023simultaneous,sangwoon2022activeextrinsic,bauza2023tac2pose} that attempt to actively consider contact in the motion model to drive robot arms to accomplish complex tasks. These methods, independently and coincidentally, employ complementarity constraints as a concise and elegant form to describe contact behaviors. Although complementarity constraints introduce additional complexity to the trajectory planning problem, they eliminate a prior specification of a mode schedule \cite{posa2014direct}. Following this formalization, we introduce complementarity constraints into the motion planning of mobile robots, enabling desired contact between the robot and movable objects.

\subsection{Feasible Trajectory Optimization of Robots}
\label{sec-related_feasible_traj}
In the contact-aware motion planning problem for robots, the feasibility of trajectories is reflected in two aspects: the consistency of states with system dynamics and the compliance of robot-environment interaction with the physical laws. 
The system dynamics is often represented as equality constraints in nonlinear programming, approximating the relationship between adjacent discrete system states \cite{russ2023underactuated}. For instance, in direct collocation methods, system dynamics are imposed with derivative constraints at collocation points \cite{hargraves1987direct,kelly2017trajecotryopt}. Alternatively, the system dynamics equation can be utilized to simulate the evolution of the robot starting from a given initial state \cite{betts1998survey}. This approach requires high-performance numerical integration algorithms and often employs error control methods to ensure the accuracy of the simulation \cite{russ2023underactuated}.

Most trajectory planning methods assume that contact between the robot and objects is prohibited. However, it is a highly conservative assumption regarding the feasibility of trajectories, as it implies that there is no need to consider the physical laws of contact \cite{wang2022gcopter,tobia2023gcs,jesus2022faster,zhou2019fastplaner}. Under this assumption, the mobile robot and obstacles are abstractly represented as combinations of convex \cite{gilbert1988fast,zhang2021obca} or non-convex \cite{zhang2023continuousSDF} geometric shapes. By detecting collisions between these geometric entities, the robot can be constrained to avoid collisions with obstacles. Additionally, we can enforce collision-free motion by constraining the robot's trajectory within a virtual safety corridor \cite{wang2022gcopter,liu2017quadSFC}. This assumption is reasonable for some applications, such as cruising or racing \cite{song2023racingSC,kaufmann2023racingNature}. However, contact between the robot and objects is inevitable or even necessary for various tasks, such as coverage planning \cite{Galceran2013corveragesurvey}, or heavy boxes pushing \cite{mataric1995cooppushing,ohashi2016realization}. Therefore, a more general contact-aware motion planning method is necessary to explore the potential locomotion of mobile robots further. 

The main challenges for the compliance of contact between robots and movable objects lie in the fidelity of the complex friction model and the accurate determination of the contact state. In the study of kinematics with the involvement of friction, the quasi-static approximation assumes that frictional contact force dominates at low velocities and that inertial forces do not play a decisive role in the motion. 
The limit surface, proposed by S. Goyal et al \cite{goyal1991planar}. is a geometric representation for mapping the friction force acting on an object to the resulting velocity. Hogan et al. \cite{hogan2020feedback} systematically analyze the kinematics of the pusher-slider system based on the concepts of quasi-static assumptions, the limit surface, and the motion cone, and propose an efficient linearization strategy to improve the computational efficiency. In addition, for complex robotic systems with multiple motion modes, Screenath et al. \cite{sreenath2013trajectory} defined transition maps to determine which mode a given system state is in. Characterizing the state of a system with multiple motion modes using a form of complementary constraints is another elegant way. 
In this work, we follow the concepts of quasi-static assumptions, limit surface, etc. We combine them with nonlinear complementarity constraints to comprehensively model the contact behavior and propose a road map to solve such a complex optimization problem.

\begin{figure*}[th]
	\centering
	\includegraphics[width=1.0\textwidth]{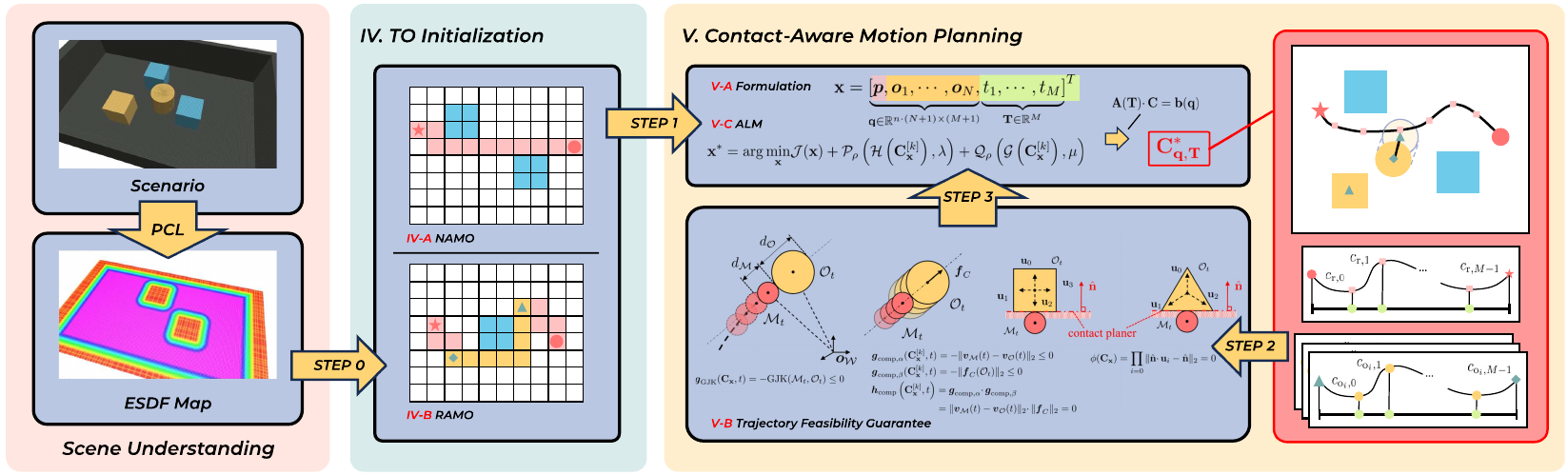}
	\caption{The overview of the CAMP framework. The figure illustrates the workflow of CAMP. The perception module of the robot provides scene understanding of the CAMP framework. In the front-end of CAMP, the appropriate searching method is selected based on specific tasks. The method calculates a sequence of states and appropriate time allocations. The back-end optimization, using the ALM as the backbone, first solves an unconstrained optimization problem and then computes energy-optimal continuous polynomials based on the local optimal decision variables. The dual variables and penalty parameters are updated based on violating feasibility constraints. These steps are repeated until feasible and optimal trajectories are obtained.}
	\label{fig-system_overview}
\end{figure*}

\section{Preliminary}
In this section, we introduce the preliminary including the fundamental spatial-temporal optimal trajectory planning problem formalization (see Section \ref{sec-spatial_temporal_opt}) and the practical ALM framework for solving the problem (see Section \ref{sec-pre_alm}).
\subsection{Spatial-Temporal Trajectory Optimization}
\label{sec-spatial_temporal_opt}
In the context, trajectory planning is considered to be the problem of finding a mapping from a one-dimensional scalar time variable $t$ to the configuration space $\mathcal{S}_{\text{config}} = \{\mathbf{q}\;|\;\mathbf{q} \in \mathbb{R}^s\}$ for a physical robotic system. The planning algorithms find a suitable function $\mathbf{C}(t): \mathbb{R} \to \mathbb{R}^s$, typically a piecewise function, to represent the trajectory of the robot. 

The spatial-temporal optimal trajectory planning, which is one of the pseudo-spectral methods \cite{russ2023underactuated,ross2012review} or sometimes known as "orthogonal collocation" \cite{russ2023underactuated,kelly2017trajecotryopt}, composed of two phases: the trajectory generation phase and the spatial-temporal numerical optimization phase, as shown in Fig.\ref{fig-system_overview}. In the former, time allocation $\mathbf{T}=[t_{1},t_{2},\cdots,t_{M}]$ and sequence of states $\mathbf{q}=[\boldsymbol{q}_{0},\boldsymbol{q}_{1},\cdots,\boldsymbol{q}_{M}]$ of trajectories composed of $M$ segments are given. Therefore, we can determine the derivative and continuity constraints that each segment needs to satisfy. Then, we solve a group of linear equations to obtain trajectory parameters that satisfy these hard constraints. In the numerical optimization phase, we simultaneously optimize the sequence of states and the time allocation while satisfying a set of complex feasibility constraints.

We consider the following optimization problem.
\begin{equation}
	\label{eq-opt_problem}
	\begin{aligned}
		\underset{\mathbf{q} ,\mathbf{T}}{\min} \thinspace  & \mathcal{J}(\mathbf{q} ,\mathbf{T})\\
		\text{subject to} \thinspace  & \mathcal{G}\left(\mathbf{C}_{\mathbf{q} ,\mathbf{T}}^{[ k]}( t)\right) \leq \mathbf{0} ,\\
		& \mathcal{H}\left(\mathbf{C}_{\mathbf{q} ,\mathbf{T}}^{[ k]}( t)\right) =\mathbf{0} ,\\
		& t\in [ 0,\| \mathbf{T} \| ] ,
	\end{aligned}
\end{equation}
where $\mathcal{J}$ is the objective function. 
The decision variables consists of the sequence of states $\mathbf{q}\in\mathbb{R}^{s\times(M+1)}$ and the time allocation $\mathbf{T}\in\mathbb{R}_{+}^{M}$. 
The inequality constraints, represented by $\mathcal{G}$, and the equality constraints, represented by $\mathcal{H}$, define the constraints that a feasible trajectory must satisfy. 

In the trajectory generation phase, a set of MINCO trajectories \cite{wang2022gcopter}, denoted by $\mathbf{C_{q,T}}$, is produced given a set of $\mathbf{q}$ and $\mathbf{T}$ by solving a quadratic programming problem with hard constraints. Details on trajectory generation can be found in [Appendix].
We can check the points which are obtained by densely sampling the generated trajectory to estimate the violation of constraints.
It is worth noting that the constraints often involve the first $k$ derivatives of the robot states, denoted by $\mathbf{C^{[k]}_{q,T}}=[\mathbf{C_{q,T}},\mathbf{\dot{C}_{q,T}},\cdots,\mathbf{C^{(k)}_{q,T}}]^{T}$, where $\mathbf{C^{(k)}_{q,T}}$ represents the $k$th derivative of the trajectory.
Therefore, we can use the chain rule to compute the partial derivatives of the constraint violation with respect to the decision variables, and then use numerical methods to find the optimal solution.

There are mainly four advantages to using this formulation: 
\begin{itemize}
	\item By directly considering the sequence of states and the time allocation as decision variables, the spatial and temporal parameters of the trajectory are simultaneously optimized. It ensures global or local optimization while maintaining trajectory continuity and consistency.
	\item We can quickly determine a parametric set of continuous trajectories for a given set of decision variables by solving the linear equations.
	\item By densely sampling points in the trajectories, we can directly check the feasibility of the trajectory, i.e., whether it satisfies the constraints.
	\item We are solving a complex constrained optimization problem, meaning that mathematical optimization theories and acceleration techniques can be leveraged to solve the problem effectively.
\end{itemize}

In this paper, we follow the classic framework of mobile robots motion planning, which divides the problem into a front-end path finding and a back-end optimization \cite{quan2020survey}. 
While the algorithms may have minor adjustments based on the types of tasks, the workflow is generally consistent. 
First, using the searching method, we find a set of feasible paths for agents based on task goals and the movable object information (see Section \ref{sec-path_searching}). 
Then, we allocate rational times for each segment of the paths. Given the time allocation, we analytically calculate continuous trajectories with minimum energy consumption while satisfying the waypoints as hard constraints. 
Finally, we use a state sequence sampled on the trajectories as the initial value for the back-end optimization (see Section \ref{sec-traj_optimization}).
\subsection{The Augmented Lagrangian Method}
\label{sec-pre_alm}
The ALM is a fundamental approach for solving constrained optimization problems. 
In CAMP, we employ the classic ALM as the backbone to solve our trajectory optimization problem. 
The general process of ALM typically consists of three steps: (i) solving an unconstrained sub-problem, (ii) analytically updating the dual variables and algorithm parameters, and (iii) checking the feasibility and optimality conditions. The corresponding augmented Lagrangian objective function can be formulated as follows for the constrained optimization problem (\ref{eq-opt_problem}).
\begin{equation}
	\label{eq-alm_obj}
	\begin{aligned}
		\mathcal{L}_{\rho }(\mathbf{x} ,\mathbf{\lambda } ,\mathbf{\mu }) =\mathcal{J}(\mathbf{x}) & +\mathcal{P}_{\rho}\left(\mathcal{H}\left(\mathbf{C}_{\mathbf{x}}^{[ k]}\right) ,\mathbf{\lambda }\right)\\
		& +\mathcal{Q}_{\rho}\left(\mathcal{G}\left(\mathbf{C}_{\mathbf{x}}^{[ k]}\right) ,\mathbf{\mu }\right),
	\end{aligned}
\end{equation}
where decision variables are defined as a vector $\mathbf{x}=[\mathbf{q},\mathbf{T}]^{T} \in \mathbb{R}^{s\cdot(M+1)+M}$, a scalar $\rho$ represents a penalty parameter of ALM, vectors $\mathbf{\lambda}$ and $\mathbf{\mu}$ are dual variables, 
and augmented terms 
$\mathcal{P}_{\rho}$ and $\mathcal{Q}_{\rho}$ 
are given by
\begin{gather}
	\mathcal{P}_{\rho}(\mathcal{H}(\mathbf{C}_{\mathbf{x}}^{[ k]}) ,\mathbf{\lambda }) =\sum _{h_i\in \mathcal{H}}\frac{\rho }{2}\left[ h_{i}\left(\mathbf{C}_{\mathbf{x}}^{[ k]}\right) +\frac{\lambda _{i}}{\rho }\right]^{2} , \label{eq-alm_eq_update}\\
	\mathcal{Q}_{\rho}(\mathcal{G}(\mathbf{C}_{\mathbf{x}}^{[ k]}) ,\mathbf{\mu }) =\sum _{g_j\in \mathcal{G}}\frac{\rho }{2}\left[\max\left( g_{j}\left(\mathbf{C}_{\mathbf{x}}^{[ k]}\right) +\frac{\mu _{j}}{\rho }\right) ,0\right]^{2} .\label{eq-alm_ieq_update}
\end{gather}

In the first step of ALM, we utilize specific unconstrained optimization methods, in detail in Section \ref{sec-opt_alm}, to obtain the global or local optimal solution $\mathbf{x}^{*}_{k}$ for the sub-problem with \ref{eq-alm_obj} as the objective function. Then, we compute the new approximate values of the Lagrangian multiples $\mathbf{\lambda }^{k} =\mathbf{\lambda }^{k} +\rho _{k}\mathcal{H}\left(\mathbf{C}_{\mathbf{x}^{*}}^{[ k]}\right)$ and $\mathbf{\mu }^{k} =\max\left(\mathbf{\mu }^{k} +\rho _{k}\mathcal{G}\left(\mathbf{C}_{\mathbf{x}^{*}}^{[ k]}\right)\right)$. The penalty parameter $\rho$ can be updated based on several strategies \cite{birgin2014practical}. 
We set $k=k+1$ and repeat the above steps until $\mathbf{x}^{*}_{k}$ satisfies the following convergence and optimality conditions.
\begin{equation*}
	\begin{aligned}
		C_{\text{opt}} : & \| \nabla \mathcal{L}_{\rho } \| _{\infty } \leq \epsilon _{\text{opt}} ,\\
		C_{\text{feas}} : & \max\left\{\begin{array}{ c }
			\underset{h_{i} \in \mathcal{H}}{\max}\left\{|h_{i}\left(\mathbf{C}_{\mathbf{x}}^{[ k]}\right) |\right\} ,\\
			\underset{g_{i} \in \mathcal{G}}{\max}\left\{g_{i}\left(\mathbf{C}_{\mathbf{x}}^{[ k]}\right)_{+}\right\},
		\end{array}\right\} \leq \epsilon _{\text{feas}} ,
	\end{aligned}
\end{equation*}
where scales $\epsilon _{\text{opt}}$ and $\epsilon _{\text{feas}}$ are user-defined thresholds.

\section{Task-Oriented Multi-Agents Path Searching}
\label{sec-path_searching}
Unlike conventional planning tasks that consider everything in the environment as obstacles, we treat all rooted entities as static obstacles while define both movable objects and the robotic systems as agents in contact-aware trajectory generation. Therefore, in the front-end of our CAMP framework, we makes full use of multi-agents path searching methods.
In this section, we introduce the specific details of our search strategy in the two fundamental tasks: the NAMO (see Section \ref{sec-searching_namo}) and the RAMO (see Section \ref{sec-searching_ramo}).
\subsection{Path Finding for Navigation Among Movable Objects}
\label{sec-searching_namo}
In a NAMO task, the prior information includes a semantic map $\mathcal{M}$ (grid map
for 2D or voxel map for 3D), the initial state of the autonomous system $\boldsymbol{s}_{0}$, and the target state of the system $\boldsymbol{s}_{t}$.The initial states of movable objects are encoded in $\mathcal{M}$ while the goal states of them are arbitrary.

Given the prior knowledge, the specific details of the path searching approach are as follows.
First, we mark the semantic labels of all movable objects in the semantic map as "free" to obtain a masked semantic map $\mathcal{M}_{\text{mask}}$. 
Then, using classical search algorithms such as A*, we can generate a collision-free path $\mathbf{p}_{\text{rbt}}=[\mathbf{s}_{0}, \mathbf{s}_{1}, \cdots, \mathbf{s}_{M}], \text{s.t.} \mathbf{s}_{M}=\mathbf{s}_{t}$ for the system within $\mathcal{M}_{\text{mask}}$. 
Finally, we associate each point $\mathbf{s}_{i}, i\in\{0,1,\cdots,M\}$ along the feasible path of the system to the initial state of the movable objects $\mathbf{o}^{\text{init}}$. 
Therefore, the state of all agents is $\displaystyle \mathbf{x}[ i] =\left[\mathbf{p}_{\text{rbt}}^{T}[ i] ,\mathbf{o}{_{1}^{\text{init}}}^{T} ,\cdots ,\mathbf{o}{_{n}^{\text{init}}}^{T}\right]^{T} ,i\in \{0,1,\cdots M\}$, where $n$ is the number of objects, and $M$ is the number of points along the path.

This approach has mainly two advantages: first, it simplifies the map, thus improving the efficiency of the search algorithm; second, it maximizes the robot's reachable space. It is important to note that contact between the robot and movable objects is not considered here.
\subsection{Path Finding for Rearrangement of Movable Objects}
\label{sec-searching_ramo}
In the RAMO task, the prior information includes all the prior information mentioned in Section \ref{sec-searching_namo} and the target states of all movable objects $\mathbf{o}^{\text{target}}$. In addition to obtaining the masked semantic map $\mathcal{M}_{\text{mask}}$, we must compute a set of action sequences $\mathcal{A}$ for all movable objects based on some decision algorithms. This set of action sequences $\mathcal{A}$ answers two questions: which movable objects will make contact with the system, and in what order will the contacts occur? Subsequently, we perform segmented path searches based on the order of contacts to find a series of collision-free paths $\mathbf{p}_{\text{rbt}}$ that connect the start and end points of each segment for the system. The following segmented discrete function defines the corresponding path of movable objects:
\begin{equation}
	\mathbf{o}_{j}[ i] =\begin{cases}
		\mathbf{o}_{j}^{\text{init}} , & i\in [ 0,p_{j}] ,\\
		\mathbf{p}_{\text{rbt}}[ i] + \delta, & i\in [ p_{j} ,\ q_{j}] ,\\
		\mathbf{o}_{j}^{\text{target}} & i\in [ q_{j} ,M] ,
	\end{cases}
\end{equation}
where $j\in \{1,2\cdots ,n\}$, $\delta$ is the position offset, $p_{j}$ and $q_{j}$ are the start and end points of the contact between the $j$-th object and the system, respectively. The state of all agents is $\displaystyle \mathbf{x}[ i] =\left[\mathbf{p}_{\text{rbt}}^{T}[ i] ,\mathbf{o}{_{1}^{T}[ i]} ,\cdots ,\mathbf{o}{_{n}^{T}[ i]}\right]^{T} ,i\in \{0,1,\cdots M\}$.
\begin{figure}[t]
	\centering
	\includegraphics[width=0.5\textwidth]{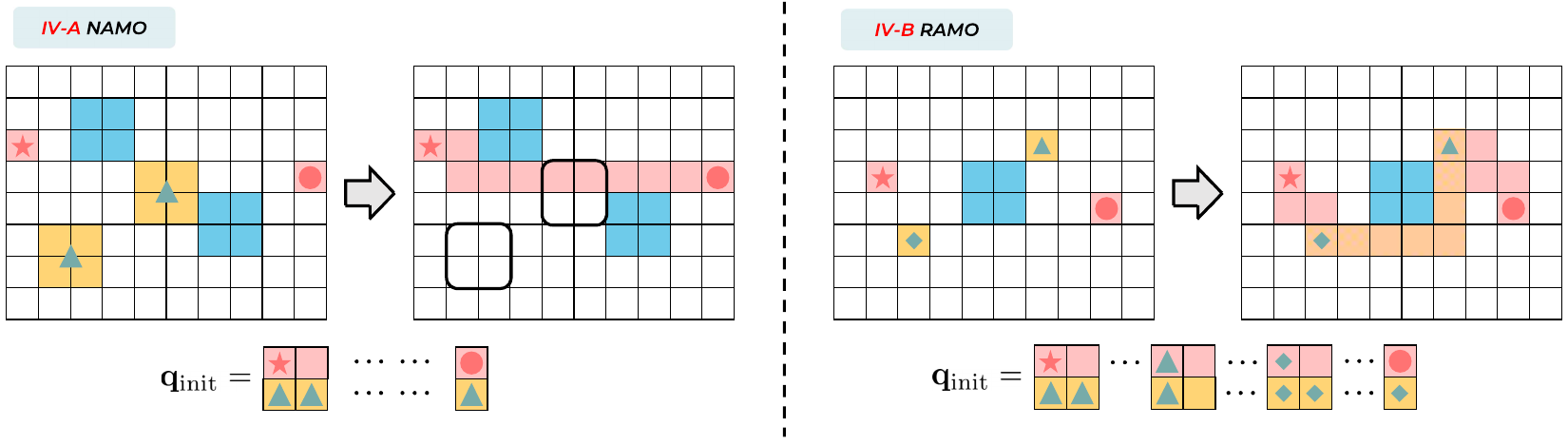}
	\caption{
		The figure illustrates front-end computations in CAMP for NAMO and RAMO tasks, showing the search method finding feasible paths for the robot in the grid map and segmented path finding for RAMO. Red denotes the robot's start and target points, while green represents  RAMO's start and target points. Light red cells indicate the robot's feasible path and light yellow cells represent feasible paths for movable objects.
	}
	\label{fig-frontend}
\end{figure}

\section{Contact-Aware Motion Planning}
\label{sec-traj_optimization}
In this section, we begin by introducing the problem formulation (see Section \ref{sec-formulation}), which includes the specific form of the objective function and the definition of the decision variables, as shown in Fig.\ref{fig-system_overview}. 
Then, we clarify how the constraints play a crucial role in the trajectory feasibility (see Section \ref{sec-constraints}).
Finally, we explain of how we solve this complex nonlinear optimization problem based on the ALM (see Section \ref{sec-opt_alm}).
\subsection{Problem Formulation}
\label{sec-formulation}
In contact-aware motion planning, we consider the following optimization problem.
\begin{equation}
	\begin{aligned}
		\text{Find} \thinspace  & \begin{array}{ l r }
			\mathbf{x} =\left[\underbrace{\boldsymbol{p} ,\boldsymbol{o}_{1} ,\cdots ,\boldsymbol{o}_{N}}_{\mathbf{q} \in \mathbb{R}^{n\cdot(N+1)\times(M+1)}} ,\underbrace{t_{1} ,\cdots ,t_{M}}_{\mathbf{T} \in \mathbb{R}^{M}}\right]^{T} & 
			\left(\begin{array}{ c }
				\text{decision}\\
				\text{variables}
			\end{array}\right)
		\end{array}\\
		\min \thinspace  & \int _{0}^{\| \mathbf{T} \| }\mathbf{C}{_{\mathbf{x}}^{( k)}}^{T}( \tau )\mathbf{WC}_{\mathbf{x}}^{( k)}( \tau ) d\tau +\boldsymbol{\alpha } \| \mathbf{T} \| +\boldsymbol{\beta }\mathcal{K}(\mathbf{C}_{\mathbf{x}} ,\mathbf{x}) ,\\
		\begin{array}{ c }
			\text{s.t.}\\
			\\\\
			\\\\
			\\\\
			\\
		\end{array} \thinspace  & \begin{array}{ l r }
			\boldsymbol{h}_{\text{bd}}(\mathbf{x}) =\mathbf{0} , & \text{(boundary conditions, Sec.\ref{sec-boundary_conditions})}\\
			\boldsymbol{g}_{\text{dist}}(\mathbf{C}^{[k]}_{\mathbf{x}} ,t) \leq \mathbf{0} , & \text{(collision avoidance, Sec.\ref{sec-collision_avoidance})}\\
			\begin{drcases}
				\boldsymbol{g}_{v}(\mathbf{C}^{[k]}_{\mathbf{x}} ,t) \leq \mathbf{0} ,\\
				\boldsymbol{g}_{a}(\mathbf{C}^{[k]}_{\mathbf{x}} ,t) \leq \mathbf{0} ,
			\end{drcases} & \text{(state bounds, Sec.\ref{sec-state_bounds})}\\
			\boldsymbol{g}_{\text{comp}}(\mathbf{C}^{[k]}_{\mathbf{x}} ,t) \leq \mathbf{0} , & \text{(contact limitations, Sec.\ref{sec-contact_limitations})}\\
			\boldsymbol{h}_{\text{dyn}}(\mathbf{C}^{[k]}_{\mathbf{x}} ,t) =\mathbf{0} , & \text{(system dynamics, Sec.\ref{sec-system_dynamics})}\\
			\boldsymbol{h}_{\text{comp}}\left(\mathbf{C}_{\mathbf{x}}^{[k]} ,t\right) =\mathbf{0} , & \text{(contact limitations, Sec.\ref{sec-contact_limitations})}\\
			\forall t\in [ 0,\| \mathbf{T} \| ] , & 
		\end{array}
	\end{aligned}
\end{equation}
where $\mathbf{x}\in\mathbb{R}^{s\cdot(M+1)+M}, s=n\cdot(N+1)$ represents the decision variables including the waypoints $\mathbf{q}$ and the time allocation $\mathbf{T}$. 
The weight matrix $\mathbf{W}$, which is a diagonal block matrix, and vectors $\boldsymbol{\alpha}$ and $\boldsymbol{\beta}$ are used to balance the terms in the objective function.
The inequality and equality constraints including collision avoidance $\boldsymbol{g}_{\text{dist}}$, state bounds $\boldsymbol{g}_{v},\boldsymbol{g}_{a}$, system dynamics $\boldsymbol{h}_{\text{dyn}}$, and contact limitations $\boldsymbol{g}_{\text{comp}}$ and $\boldsymbol{h}_{\text{comp}}$.
The loss function $\mathcal{K}$ intuitively incorporates user-defined metrics into the optimization problem. For instance, we can measure the rationality of the movable objects' goal state in NAMO tasks using the loss function $\mathcal{K}$. 

We utilize polynomial functions to represent the trajectories $\mathbf{C_x}$. This means that for each agent (including the robot and the movable objects), the $i$th segment in one dimension of its configuration space can be expressed as 
\begin{equation}
	\boldsymbol{c}_{i}( t) =\sum _{j=0}^{2k-1} c_{i,j}( t-T_{i-1})^{j} ,\ \forall t\in [ T_{i-1} ,T_{i}],
\end{equation}
where
\begin{equation}
	T_{i} =T_{0} +\sum _{p=1}^{i} t_{p} ,i\in \{1,\cdots ,M\} ,T_{0} =T_{\text{init}}.
\end{equation}

\begin{remark}
	According to Theorem 2 in \cite{wang2022gcopter}, given the endpoints of each trajectory segment and the time allocation, we can naturally determine the trajectories $\mathbf{C_x}$ with minimum energy consumption. Given this foundation, we further exploit the efficiency improvement brought by dimensionality reduction through polynomial parameterization without compromising the feasibility of the trajectory.
\end{remark}
\begin{remark}
	The complementarity constraints (specific form provided in Section \ref{sec-contact_limitations}) are naturally incorporated into this formulation, making the problem an optimization problem with complementarity constraints (OPCC). Since complementarity constraints usually violate the linear independent constraint qualification (LICQ), it requires special care on the numerical solver to address such challenging \cite{izmailov2012global,howell2022calipso}. ALM can alleviate the aforementioned issues \cite{birgin2014practical,fernandez2012local}, which is one of the reasons we choose ALM as the backbone in CAMP (see Section \ref{sec-opt_alm} for more details).
\end{remark}
\begin{figure}[t]
\centering
\includegraphics[width=0.5\textwidth]{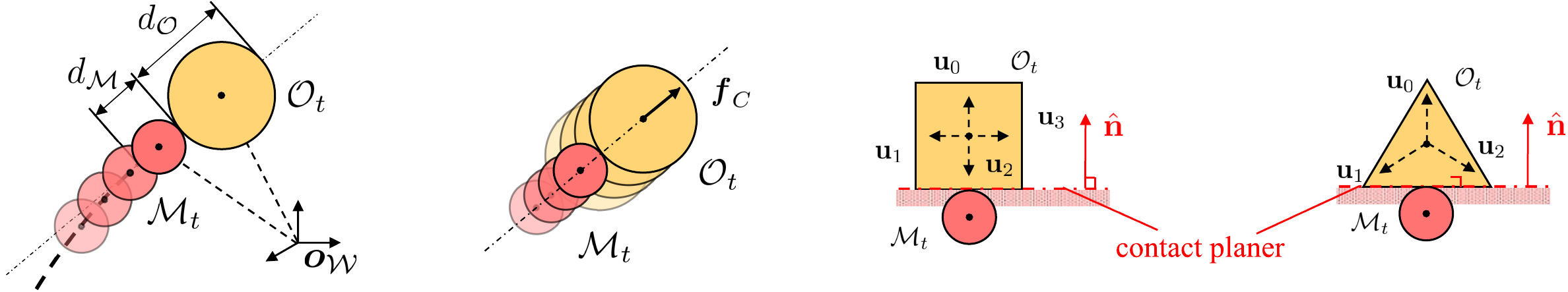}
\caption{
	The figure of the feasibility constraints. The two left subplots show the complementarity constraints for non-penetration between the robot and objects. The two right subplots illustrate the constraints on the contact force or velocity direction for two simple geometric contours in RAMO tasks. }
\label{fig-constraints}
\end{figure}
\subsection{Trajectory Feasibility Guarantees}
\label{sec-constraints}
We ensure the feasibility of the motion by checking the satisfaction of all constraints in densely sampled points along the trajectories. 
For each node, constraint satisfaction is evaluated. By utilizing the chain rule, we backpropagate the gradient of constraints violation concerning the trajectory coefficients to the decision variables. Numerical methods can leverage the gradient to optimize the decision variables. Therefore, while performing simultaneous spatial-temporal optimization, the feasibility of the trajectories is maximized. Next, we provide a detailed explanation of the constraints that guarantee the feasibility of the motion. 
\subsubsection{Boundary Conditions}
\label{sec-boundary_conditions}
In the NAMO tasks, the target state of the movable objects is arbitrary (but must be feasible), while in the RAMO tasks, the target state of the movable objects is predetermined and fixed. We represent the boundary conditions using equality constraints $\boldsymbol{h}_{\text{bd}}(\mathbf{x})=\mathbf{A}\mathbf{q}-\mathbf{b}=\mathbf{0}$. We use a mask matrix $\mathbf{A}\in\mathbb{R}^{s\cdot(M+1)}$ to indicate whether we are concerned about the target state of the movable objects. 
For example, in NAMO tasks, the mask value for the target constraint of the movable objects is 0.
A vector $\mathbf{b}\in\mathbb{R}^{s\cdot(M+1)}$ represents the desired states.
\subsubsection{Collision Avoidance}
\label{sec-collision_avoidance}
We consider two primary collisions: (i) collisions between agents and collisions between agents and the static obstacles. We utilize the Gilbert-Johnson-Keerthi (GJK) method \cite{gilbert1988fast} to detect collisions between agents, represented by $\displaystyle \boldsymbol{g}_{\text{GJK}}(\mathbf{C}_{\mathbf{x}} ,t) =-\textrm{GJK}(\mathcal{A}_{t} ,\mathcal{B}_{t}) \leq \mathbf{0}$, where $\mathcal{A}$ and $\mathcal{B}$ are different agents represented as convex hulls. To simplify the problem complexity, we treat the robot as a circle (when $n=2$) or a sphere (when $n=3$), represented by $\displaystyle \mathcal{M}(\boldsymbol{p}( t) ,r) =\left\{\boldsymbol{p}_{r} \in \mathbb{R}^{n} \thinspace |\thinspace \| \boldsymbol{p}_{r} -\boldsymbol{p} \| _{2} \leq r\right\}$. The movable objects, on the other hand, are treated as convex polygons or convex hulls, represented by $\displaystyle \mathcal{O}(\boldsymbol{o}_{i}( t)) =\left\{\boldsymbol{o} \in \mathbb{R}^{n} \thinspace |\thinspace \mathbf{A} \cdot (\boldsymbol{o} -\boldsymbol{o}_{i}) \leq \mathbf{b}\right\}$. Collision avoidance of agents in the static obstacles is achieved by ensuring the value of the ESDF is non-negative, denoted as $\displaystyle \boldsymbol{g}_{\text{ESDF}}(\mathbf{C}_{\mathbf{x}} ,t) =-\textrm{ESDF}(\mathbf{C}_{\mathbf{x}} ,t) \leq \mathbf{0}$.
\subsubsection{State Bounds}
\label{sec-state_bounds}
To ensure the safety of robot motion, we impose constraints on the maximum velocity and maximum acceleration of the agents, given by $\displaystyle \boldsymbol{g}_{v}(\mathbf{C}^{[k]}_{\mathbf{x}} ,t) =\Vert \dot{\mathbf{C}}_{\mathbf{x}}( t)\Vert _{2}^{2} -\boldsymbol{v}_{\text{max}}^{2} \leq \mathbf{0}$ and $\displaystyle \boldsymbol{g}_{a}(\mathbf{C}^{[k]}_{\mathbf{x}} ,t) =\Vert \ddot{\mathbf{C}}_{\mathbf{x}}( t)\Vert _{2}^{2} -\boldsymbol{a}_{\text{max}}^{2} \leq \mathbf{0}$, respectively.
\subsubsection{System Dynamics}
\label{sec-system_dynamics}
The dynamics of agents can be described by the Newton-Euler equations, given by $\displaystyle \mathbf{M} \cdot \ddot{\mathbf{C}}_{\mathbf{x}}( t) =\mathbf{f}(\mathbf{C}_{\mathbf{x}}( t) ,\dot{\mathbf{C}}_{\mathbf{x}}( t) ,t)$. As movable objects do not possess active power sources, contact forces must provide the necessary forces for their motion. Therefor, the system dynamics equation of movable objects is $\displaystyle \boldsymbol{h}_{\text{dyn}}\left(\mathbf{C}_{\mathbf{x}}^{[k]} ,t\right) =\mathbf{M} \cdot \ddot{\mathbf{C}}_{\mathbf{x}}( t) -\mathbf{J}^{T}\boldsymbol{\lambda } +\boldsymbol{f}_{D}=\mathbf{0}$, where the Jacobian $\mathbf{J}=\frac{\boldsymbol{\phi}(\mathbf{C_x})}{\partial{\mathbf{C_x}}}$ represents the non-penetration constraints $\boldsymbol{\phi}(\mathbf{C_x})$, such as gap functions \cite{sheldon2022contact}, with respect to the system state $\mathbf{C_x}$. $\boldsymbol{\lambda}$ is a vector of constraint impulse force magnitudes, and $\boldsymbol{f}_{D}$ represents air drag (which can be ignored at low speeds).
Intuitively, the contact force $\boldsymbol{f}_{C} = \mathbf{J}^{T}\boldsymbol{\lambda }$ ensures the feasibility of motion between multiple objects, such as maintaining non-penetration between objects, keeping legal directions of movement based on the geometry of the contour, as shown in Fig.\ref{fig-constraints}. It is important to note that the contact point and direction of contact force are determined during the collision detection phase and depend on the geometric characteristics of the interacting objects. For example, the contact force direction is opposite to the surface unit normal direction for smooth objects. This philosophy is widely adopted by contact simulation \cite{sheldon2022contact}.
\subsubsection{Contact Limitations}
\label{sec-contact_limitations}
One of the most important constraints in CAMP to ensure the feasibility of trajectories is to prevent agents from penetrating each other, which is represented as complementarity constraints. We make three observations between two contacting objects, $\mathcal{O}_{i}$ and $\mathcal{O}_{j}$. (i) The magnitude of their relative velocity is always non-negative, given by $\displaystyle \boldsymbol{g}_{i ,j}(\mathbf{C}^{[k]}_{\mathbf{x}} ,t) =-\| \dot{\mathcal{O}}_{i}(t) -\dot{\mathcal{O}}_{j}(t) \| _{2} \leq 0$. Similarly, (ii) the magnitude of the contact frictional force for one object is also non-negative, given by $\displaystyle \boldsymbol{g}_{k}(\mathbf{C}^{[k]}_{\mathbf{x}} ,t) =-\| \boldsymbol{f}_{C}(\ddot{\mathcal{O}}_{k}(t))\| _{2} \leq 0$. Moreover, (iii) at least one of these magnitudes is zero, given by $\displaystyle \boldsymbol{h}_{\text{comp}}\left(\mathbf{C}^{[k]}_{\mathbf{x}} ,t\right) =\boldsymbol{g}_{i ,j} \cdot \boldsymbol{g}_{k} =\| \dot{\mathcal{O}}_{i}(t) -\dot{\mathcal{O}}_{j}(t) \| _{2} \cdot \| \boldsymbol{f}_{C}(\ddot{\mathcal{O}}_{k}(t))\| _{2} =0, k\in\{i,j\}$. These (i), (ii), and (iii) conditions constitute the complementarity constraints, which are also commonly found in computer graphics for contact simulation \cite{sheldon2022contact} and in robotics for bipedal or wheeled-legged robots control \cite{vicotr2023nonsmoothcontact}. For more details on the friction dynamics of the system, please refer to Appendix .
\subsection{Optimization Using ALM}
\label{sec-opt_alm}
We utilize the ALM to obtain the contact-aware feasible trajectories efficiently. First, we introduce the dual variables $\boldsymbol{\lambda}$ and $\boldsymbol{\mu}$ to transform the optimization problem mentioned in Section \ref{sec-formulation} into an augmented Lagrangian form based on Equation \ref{eq-alm_obj}, \ref{eq-alm_eq_update}, and \ref{eq-alm_ieq_update}. Next, we utilize the initial values provided by the front-end searching method and solve the unconstrained sub-problem with the augmented Lagrangian function as the objective function. To solve this sub-problem efficiently, we employ a classic quasi-Newton numerical method. Specifically, we use the L-BFGS method \cite{liu1989limited} to determine the appropriate gradient descent direction and the Lewis-Overton method \cite{lewis2013nonsmooth} as a line search method to find the suitable step length. Afterward, we analytically update the dual variables $\boldsymbol{\lambda}$, $\boldsymbol{\mu}$, and the penalty parameter $\rho$ using the following formulas.
\begin{equation*}
\begin{aligned}
\mathbf{\lambda }^{k} & =\mathbf{\lambda }^{k} +\rho _{k}\mathbf{h}\left(\mathbf{C}_{\mathbf{x}^{*}}^{[ k]}\right),\\
\mathbf{\mu }^{k} & =\max\left(\mathbf{\mu }^{k} +\rho _{k}\mathbf{g}\left(\mathbf{C}_{\mathbf{x}^{*}}^{[ k]}\right) ,0\right),\\
\rho ^{k} & =\gamma \rho ^{k}, \gamma > 1.
\end{aligned}
\end{equation*}

ALM demonstrates superior performance in solving OPCC. As we discussed in Section \ref{sec-formulation}, complementarity constraints usually do not satisfy the LICQ, but the convergence analysis of ALM does not depend on LICQ \cite{birgin2014practical,fernandez2012local}. Related research \cite{izmailov2012global} shows that ALM is robust compared to many commonly used solvers for OPCC. Since ALM belongs to a class of penalty methods that eliminate constraints by adding augmented terms to the objective function \cite{bertsekas2014constrained,nocedal1999numerical}, it is possible to take full advantage of the robust and efficient unconstrained optimization methods. For readers who wish to have a more detailed introduction and analysis of ALM, we recommend referring to \cite{birgin2014practical}.

\section{Results}
In this section, we thoroughly evaluate the CAMP algorithm through simulations and experiments for the NAMO and RAMO tasks. Specifically, we aimed to address the following three questions:
\begin{itemize}
	\item[i)] Can the CAMP expand the locomotion of mobile robots compared to contact-avoidance methods, enabling them to accomplish tasks they cannot handle by using the contact-avoidance methods (see Section.\ref{sec-benchmark})?
	\item[ii)] Are the trajectories generated by the CAMP feasible, particularly in long-duration, long-distance RAMO tasks that require active contact?
	\item[iii)] Can the CAMP be easily deployed in different tasks when there are changes in user-defined task objectives?
\end{itemize}

Through results analysis, we sought to provide answers to these questions and evaluate the effectiveness and versatility of the CAMP.
\begin{table*}[]
	\linespread{1.25}
	\scriptsize
	\caption{Comparison of time consumption and success rates between the CAMP and the baseline algorithm in NAMO and RAMO tasks.}
	\label{tab:benchmark}
	\begin{tabular}{cc|ccc|cccccccc}
		\hline
		\multicolumn{2}{c|}{\multirow{2}{*}{Algorithms}}           & \multicolumn{3}{c|}{\multirow{2}{*}{Front-end}}                                             & \multicolumn{8}{c}{Back-end}                                                                                                                                                                                                                                                                                                                 \\ \cline{6-13} 
		\multicolumn{2}{c|}{}                                      & \multicolumn{3}{c|}{}                                                                       & \multicolumn{4}{c|}{Ours}                                                                                                                                                       & \multicolumn{4}{c}{GCOPTER}                                                                                                                                \\ \hline
		\multicolumn{2}{c|}{Tasks}                                 & min {[}ms{]} & max {[}ms{]} & \begin{tabular}[c]{@{}c@{}}mean (std)\\ {[}ms{]}\end{tabular} & min {[}s{]} & max {[}s{]} & \begin{tabular}[c]{@{}c@{}}mean (std)\\ {[}s{]}\end{tabular} & \multicolumn{1}{c|}{\begin{tabular}[c]{@{}c@{}}success\\ rate {[}\%{]}\end{tabular}} & min {[}s{]} & max {[}s{]} & \begin{tabular}[c]{@{}c@{}}mean (std)\\ {[}s{]}\end{tabular} & \begin{tabular}[c]{@{}c@{}}success\\ rate {[}\%{]}\end{tabular} \\ \hline
		\multicolumn{1}{c|}{\multirow{8}{*}{NAMO}} & S7M1          & 1.68         & 78.78        & 21.71 (19.93)                                                 & 1.08        & 4.08        & 2.45 (0.89)                                                  & \multicolumn{1}{c|}{75}                                                              & 0.22        & 0.51        & 0.38 (0.09)                                                  & 55                                                              \\
		\multicolumn{1}{c|}{}                      & S6M2          & 2.35         & 35.32        & 13.42 (10.23)                                                 & 1.05        & 5.51        & 3.07 (1.22)                                                  & \multicolumn{1}{c|}{90}                                                              & 0.15        & 0.91        & 0.45 (0.21)                                                  & 55                                                              \\
		\multicolumn{1}{c|}{}                      & S5M3          & 2.29         & 41.72        & 11.19 (10.44)                                                 & 1.47        & 9.60        & 4.37 (2.05)                                                  & \multicolumn{1}{c|}{100}                                                             & 0.30        & 0.91        & 0.57 (0.19)                                                  & 50                                                              \\
		\multicolumn{1}{c|}{}                      & S4M4          & 1.66         & 72.49        & 16.98 (19.85)                                                 & 2.74        & 10.18       & 4.90 (2.05)                                                  & \multicolumn{1}{c|}{100}                                                             & 0.19        & 0.89        & 0.58 (0.16)                                                  & 60                                                              \\
		\multicolumn{1}{c|}{}                      & S3M5          & 2.08         & 43.85        & 9.11 (11.90)                                                  & 2.71        & 11.78       & 7.06 (2.83)                                                  & \multicolumn{1}{c|}{100}                                                             & 0.26        & 0.87        & 0.56 (0.21)                                                  & 45                                                              \\
		\multicolumn{1}{c|}{}                      & S2M6          & 1.58         & 22.26        & 6.37 (5.98)                                                   & 4.19        & 15.07       & 7.75 (2.91)                                                  & \multicolumn{1}{c|}{100}                                                             & 0.43        & 1.13        & 0.78 (0.22)                                                  & 55                                                              \\
		\multicolumn{1}{c|}{}                      & S1M7          & 1.74         & 41.67        & 5.69 (8.64)                                                   & 4.44        & 18.02       & 9.52 (3.56)                                                  & \multicolumn{1}{c|}{100}                                                             & 0.41        & 1.64        & 0.83 (0.31)                                                  & 60                                                              \\ \cline{2-13} 
		\multicolumn{1}{c|}{}                      & TOTAL AVG     & 1.91         & 48.01        & 12.07                                                         & 2.53        & 10.61       & 5.59                                                         & \multicolumn{1}{c|}{95}                                                              & 0.28        & 0.98        & 0.59                                                         & 54.3                                                            \\ \hline
		\multicolumn{1}{c|}{\multirow{2}{*}{RAMO}} & S4M1-Cylinder & 1.20         & 23.25        & 5.90 (6.25)                                                   & 2.25        & 5.41        & 3.76 (0.91)                                                  & \multicolumn{1}{c|}{100}                                                             & \multicolumn{4}{c}{\multirow{2}{*}{----}}                                                                                                                  \\
		\multicolumn{1}{c|}{}                      & S4M1-Cubic    & 1.19         & 10.85        & 3.17 (2.24)                                                   & 1.23        & 6.24        & 3.19 (1.26)                                                  & \multicolumn{1}{c|}{95}                                                              & \multicolumn{4}{c}{}                                                                                                                                       \\ \hline
	\end{tabular}
\end{table*}
\subsection{Simulation Analysis: NAMO and RAMO Tasks}
\label{sec-benchmark}
To address the question i), we design experiments to validate the CAMP algorithm's superior navigation and rearrangement capability compared to contact-avoidance trajectory planning methods in crowded scenarios. Regarding algorithm design, we follow the front-end to back-end planning framework. We use the A* algorithm to find feasible paths for the front-end path searching. In the back-end optimization, we use the GCOPTER algorithm \cite{wang2022gcopter}, which is currently a SOTA method in mobile robot motion planning, as the baseline method.
\begin{figure}[t]
	\centering
	\includegraphics[width=0.35\textwidth]{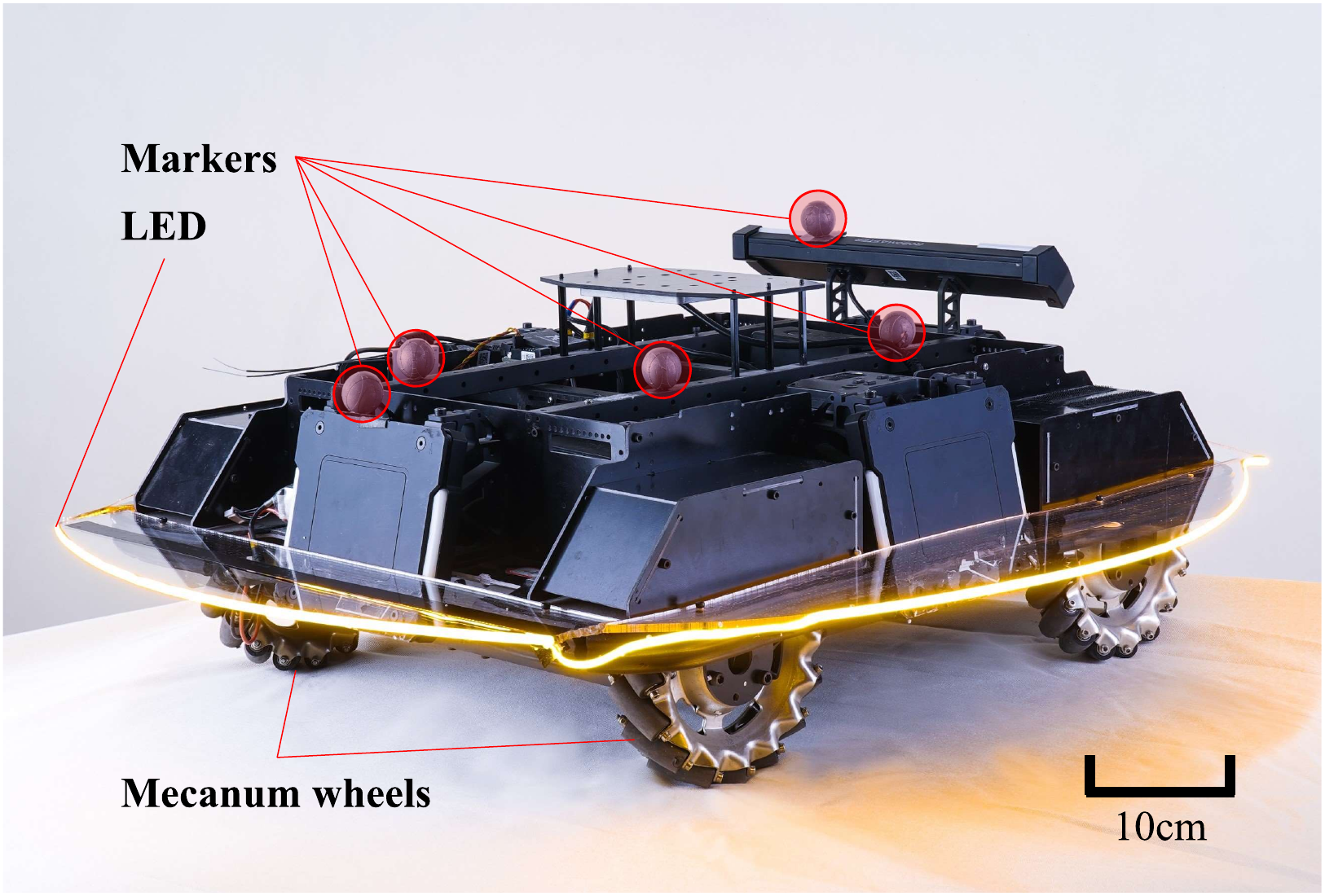}
	\caption{The omnidirectional mobile robot used in our real world experiments.}
	\label{fig-rmcar}
\end{figure}

To thoroughly evaluate the performance of the algorithm, we design NAMO tasks following specific rules (all coordinate values are in the world frame):
\begin{itemize}
	\item In all tasks, the starting point ($\boldsymbol{p}_{s}=[0.0, 0.0]^T$) and the target point ($\boldsymbol{p}_{t}=[6.0, 0.0]^T$) of the robot are fixed.
	\item The number of movable objects and static obstacles in all tasks is constant, totaling eight objects. The object's geometric center $\boldsymbol{o}=[x, y]^T$ are randomly generated, where $\boldsymbol{o}$ is limited within th space $\mathcal{S}=\{\boldsymbol{o}=[x, y]^T, x\in[1.0, 5.0], y\in[-2.0, 2.0]\}$.
	\item Assuming the number of movable objects is $m$, we generate 7 sets of tasks, with $m$ range from 1 to 7. We randomly generate 20 maps per set.
	\item We evaluate the CAMP and baseline method on the same maps and record optimization time, task success rate, and other relevant data.
\end{itemize}
Furthermore, we design a set of RAMO tasks as benchmarks. Compared to the NAMO tasks, there are several changes:
\begin{itemize}
	\item The total number of static obstacles is fixed at 4. The obstacles' geometric center $\boldsymbol{o}=[x, y]^T$ are randomly generated, where $\boldsymbol{o}$ is limited within th space $\mathcal{S}=\{\boldsymbol{o}=[x, y]^T, x\in[2.0, 4.0], y\in[-2.0, 2.0]\}$.
	\item The starting point and target point of the movable object are $\boldsymbol{o}_s=[1.0, y_{s}]^T$ and $\boldsymbol{o}_t=[5.0, y_{t}]^T$, respectively. The coordinate $y_{s}$ and $y_{t}$ are randomly generated and are limited within $[-2.0, 2.0]$.
	\item The geometric shapes of the movable objects are divided into two types: cylinder and cube. The cylinder can move in any direction, while the cubic one can only move along the x- and y-axis of the world frame.
\end{itemize}

Table \ref{tab:benchmark} presents the time consumption and success rates of our method and the baseline algorithm in each group of tasks. We observe two facts based on the data related to the NAMO tasks. First, in all groups of tasks, our proposed CAMP method achieves significantly higher success rates than the contact-avoidance approach. Second, as the number of movable objects in the scene increases, the success rate of our method also increases (with a success rate exceeding 90\% when the number of movable objects exceeds two). These observations indicate that our method can leverage the property of movable objects in the scenarios, expanding the reachable space for mobile robots and significantly improving the success rate of navigation tasks. The data related to the RAMO tasks in the table demonstrates the outstanding success rate of our method with limited optimization time (more details about the simulations can be found in the video 
 of supplementary materials).
\begin{table}[]
	\linespread{1.25}
	\scriptsize
	\caption{Tracking error of real-world experiments.}
	\label{tab:tracking-error}
	\begin{tabular}{ccc|cc|cc}
		\hline
		\multicolumn{3}{c|}{\multirow{2}{*}{Task}}                                                                                                                          & \multicolumn{2}{c|}{Scenario 1 (NAMO)}                                       & \multicolumn{2}{c}{Scenario 2 (RAMO)} \\ \cline{4-7} 
		\multicolumn{3}{c|}{}                                                                                                                                               & \begin{tabular}[c]{@{}c@{}}GCOPTER\\ Method\end{tabular} & CAMP   & Cylinder        & Cube         \\ \hline
		\multicolumn{1}{c|}{\multirow{8}{*}{Robot}}                                                    & \multicolumn{1}{c|}{\multirow{4}{*}{POS}} & MEAN {[}m{]}      & 0.0167                                                       & 0.0164 & 0.0149          & 0.0207       \\
		\multicolumn{1}{c|}{}                                                                          & \multicolumn{1}{c|}{}                          & STD {[}m{]}       & 0.0088                                                       & 0.0081 & 0.0106          & 0.0093       \\
		\multicolumn{1}{c|}{}                                                                          & \multicolumn{1}{c|}{}                          & MAX {[}m{]}       & 0.0425                                                       & 0.0391 & 0.0718          & 0.0725       \\
		\multicolumn{1}{c|}{}                                                                          & \multicolumn{1}{c|}{}                          & MIN {[}m{]}       & 0.0019                                                       & 0.0012 & 0.0001          & 0.0001       \\ \cline{2-7} 
		\multicolumn{1}{c|}{}                                                                          & \multicolumn{1}{c|}{\multirow{4}{*}{VEL}} & MEAN {[}m/s{]}    & 0.0409                                                       & 0.0385 & 0.0387          & 0.0776       \\
		\multicolumn{1}{c|}{}                                                                          & \multicolumn{1}{c|}{}                          & STD {[}m/s{]}     & 0.0288                                                       & 0.0223 & 0.0369          & 0.0609       \\
		\multicolumn{1}{c|}{}                                                                          & \multicolumn{1}{c|}{}                          & MAX {[}m/s{]} & 0.2015                                                       & 0.1107 & 0.2077          & 0.3201       \\
		\multicolumn{1}{c|}{}                                                                          & \multicolumn{1}{c|}{}                          & MIN {[}m/s{]} & 0.0007                                                       & 0.0006 & 0.0007          & 0.0012       \\ \hline
		\multicolumn{1}{c|}{\multirow{8}{*}{\begin{tabular}[c]{@{}c@{}}Movable\\ Object\end{tabular}}} & \multicolumn{1}{c|}{\multirow{4}{*}{POS}} & MEAN {[}m{]}      & -                                                            & 0.1307 & 0.1707          & 0.2018       \\
		\multicolumn{1}{c|}{}                                                                          & \multicolumn{1}{c|}{}                          & STD {[}m{]}       & -                                                            & 0.1312 & 0.1460          & 0.1533       \\
		\multicolumn{1}{c|}{}                                                                          & \multicolumn{1}{c|}{}                          & MAX {[}m{]}       & -                                                            & 0.3125 & 0.4136          & 0.5124       \\
		\multicolumn{1}{c|}{}                                                                          & \multicolumn{1}{c|}{}                          & MIN {[}m{]}       & -                                                            & 0.0058 & 0.0040          & 0.0090       \\ \cline{2-7} 
		\multicolumn{1}{c|}{}                                                                          & \multicolumn{1}{c|}{\multirow{4}{*}{VEL}} & MEAN {[}m/s{]}    & -                                                            & 0.0411 & 0.0522          & 0.0940       \\
		\multicolumn{1}{c|}{}                                                                          & \multicolumn{1}{c|}{}                          & STD {[}m/s{]}     & -                                                            & 0.0643 & 0.0668          & 0.0882       \\
		\multicolumn{1}{c|}{}                                                                          & \multicolumn{1}{c|}{}                          & MAX {[}m/s{]}     & -                                                            & 0.3456 & 0.4473          & 0.3413       \\
		\multicolumn{1}{c|}{}                                                                          & \multicolumn{1}{c|}{}                          & MIN {[}m/s{]}     & -                                                            & 0.0001 & 0.0002          & 0.0008      
	\end{tabular}
\end{table}
\begin{figure*}[th]
	\centering
	\includegraphics[width=1.0\textwidth]{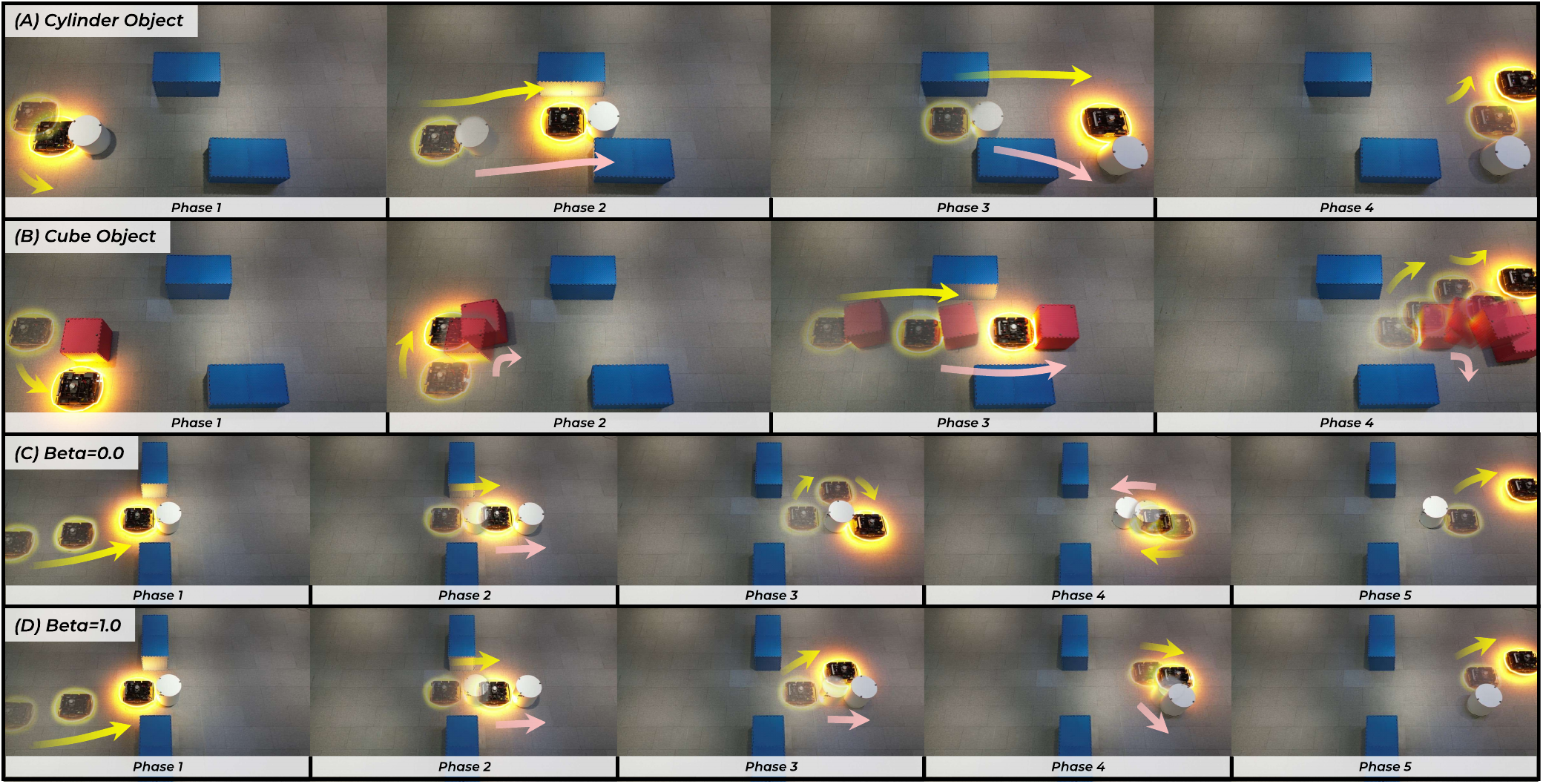}
	\caption{Experiments demonstrations. The first and second rows show a series of snapshots, illustrating the robot's execution of RAMO tasks involving cylinder and cube objects, respectively. The third and fourth rows demonstrate different robot trajectories produced given different parameters in the same NAMO scenario.}
	\label{fig-exp-tasks}
\end{figure*}
\subsection{Experiments Validation}
We conduct real-world experiments to validate the feasibility of the trajectories generated by our method, including in long-duration, long-distance, and active contact RAMO tasks to answer the question ii). The robot used in our experiments is an omnidirectional mobile robot (as shown in Fig.\ref{fig-rmcar}). To simplify the problem, we install curved acrylic panels on the robot's sides and highlight its contour with yellow LEDs. 

As shown in Fig.\ref{fig-introduction} and Fig.\ref{fig-exp-tasks} (A)-(B), we deploy our algorithm in two different scenarios. In scenario 1
(see Fig.\ref{fig-introduction}), we compared the trajectories generated by the contact-avoidance method and the CAMP. The former chose to navigate around all objects, while CAMP chose to push the movable object, resulting in shorter movement distances for the robot. In scenario 2
(see Fig.\ref{fig-exp-tasks} (A) and (B)), we perform RAMO tasks with cylindrical and cubical movable objects, respectively. The mobile robot can smoothly execute the reference trajectories and push objects with different geometric shapes to their desired positions.

We record the robot's and movable objects' actual movement using a motion capture system and calculate the trajectory tracking error (as shown in Table.\ref{tab:tracking-error}). The actual performance and tracking error demonstrate that even in long-duration, long-distance, and active contact RAMO tasks, the trajectories produced by CAMP are feasible and can be executed successfully. These experiments provide practical evidence of the feasibility of our method in generating trajectories for challenging RAMO tasks.

\subsection{Trajectories Customization}
To answer question iii) posed at the beginning of this section, we evaluate the trajectories generated by CAMP with different objective functions in the same NAMO task.

As shown in Fig.\ref{fig-exp-tasks} (C) and (D), the robot need to complete a NAMO task.
Due to the narrow passage between static obstacles occupied by a movable object, the robot must contact the object first and then move to its target position. Since the final position of the movable object in the NAMO task is arbitrary, we include an evaluation function as a term in the objective function, representing our preference for the final position of the movable objects to not overlap with the robot's trajectory. We employ  the method proposed by Lakshmanan et al. \cite{Hovakimyan-RSS-19} to calculate the minimum distance between the target position of the movable object and the robot's trajectories as the evaluation function. We set the coefficient $\beta$ of the term to 0 and 1, respectively, resulting in two different trajectories.

We find that if we do not care about the final position of the movable object, i.e., $\beta=0$, the robot tends to push the object back towards its initial position after moving it. However, if we care about the object's final position, i.e., $beta=1.0$, the robot no longer pushes the object back after moving it. These findings indicate that we can quickly deploy the algorithm to different tasks by adjusting the objective function in CAMP. For example, we can decide whether the robot should turn back and close a door after pushing it open.

This experiment demonstrates the flexibility of CAMP in generating trajectories based on different user-defined objectives, allowing us to customize the robot's behavior according to specific task requirements.

\section{Conclusion} 
\label{sec:conclusion}
In this work, we propose a novel contact-aware motion planning method for robots among movable objects. Our method incorporates the contact between the robots and objects as complementarity constraints in trajectory planning. We efficiently solve this constrained optimization problem using the ALM. Through extensive simulations, our proposed CAMP method demonstrates enhanced locomotion capabilities for robots compared to baseline approaches, significantly improving the success rates of NAMO and RAMO tasks in crowded scenarios. Furthermore, diverse real-world experiments show that the trajectories generated by our method are feasible and can be quickly deployed in various tasks.

\appendices
\section{Contact Friction Dynamics}
\label{appendix_a}
In contact-aware trajectory generation, it is essential to consider the friction between agents and the ground and the friction that occurs when agents contact each other. Following the quasi-static assumption made in many previous works \cite{hogan2020feedback}, we establish the nonlinear complementarity constraints (see Section \ref{sec-contact_limitations}) based on the Coulomb Friction Law. 
We emphasize three concepts as the same as the previous work \cite{hogan2020feedback}. Firstly, the quasi-static assumption states that friction between the robot and the movable objects dominates in low-speed movement. Therefore, we can ignore the frictional force between the objects and the ground. Secondly, the limit surface establishes a mapping from the frictional force applied on the object to the resulting velocity. Third, motion cone is used to determine the relative motion between the robot and the movable object, i.e., in sliding mode or stick mode.

\bibliographystyle{IEEEtranN}
\bibliography{References}
\end{document}